\documentclass[aip,graphicx]{revtex4-1}
\usepackage{geometry}              		% See geometry.pdf to learn the layout options. There are lots.
\usepackage[dvipdfmx]{graphicx}
\usepackage{xcolor}				% Use pdf, png, jpg, or eps§ with pdflatex; use eps in DVI mode
\usepackage[T1]{fontenc}
\usepackage{mathptmx}
\usepackage{amssymb}
\usepackage[utf8]{inputenc}
\usepackage{amsmath}
\usepackage{amsthm}
\usepackage{amsfonts}
\usepackage{bm}
\usepackage{hyperref}
\usepackage[caption=false]{subfig}
\usepackage{CJK}
\newcommand{\norm}[1]{\left\lVert#1\right\rVert}
\theoremstyle{definition}

\DeclareMathOperator*{\argmin}{arg\,min}
\usepackage{algorithm}
\usepackage{algpseudocode}
\draft % marks overfull lines with a black rule on the right

\begin{document}

%\preprint{}

\title{Robot formation control in nonlinear manifold using Koopman operator theory} %Title of paper



\author{Yanran Wang}
\email{y-wang@dove.kuee.kyoto-u.ac.jp}
% \altaffiliation

\author{Tatsuya Baba}
\author{Takashi Hikihara}%
\email{hikihara.takashi.2n@kyoto-u.ac.jp}
\affiliation{%
 Department of Electrical Engineering, Kyoto University, Kyoto, 615-8510 Japan
}%


\date{\today}

\begin{abstract}
Formation control of multi-agent systems has been a prominent research topic, spanning both theoretical and practical domains over the past two decades. Our study delves into the leader-follower framework, addressing two critical, previously overlooked aspects. Firstly, we investigate the impact of an unknown nonlinear manifold, introducing added complexity to the formation control challenge. Secondly, we address the practical constraint of limited follower sensing range, posing difficulties in accurately localizing the leader for followers. Our core objective revolves around employing Koopman operator theory and Extended Dynamic Mode Decomposition to craft a reliable prediction algorithm for the follower robot to anticipate the leader's position effectively. Our experimentation on an elliptical paraboloid manifold, utilizing two omni-directional wheeled robots, validates the prediction algorithm's effectiveness.
\end{abstract}


\maketitle %\maketitle must follow title, authors, abstract and \pacs

\section{Introduction}

In recent years, the field of multi-robot systems has witnessed remarkable advancements, particularly in the area of swarm robotics, enabling the deployment of teams of robots for a wide range of applications, including exploration, navigation, transpotation and search missions \cite{applychen, applymcguire, applypons, applyzhang, applymarjovi}. Swarm robotics, inspired by the collective behaviors observed in social insects like ants, bees, and termites, focuses on the study of decentralized and self-organized systems, where individual robots interact locally to achieve complex tasks collectively \cite{Swarm_robotics_review}. 

One of the critical aspects of swarm robotics is formation control, which involves maintaining desired spatial arrangements among the robots as they navigate through their environment. Researchers have approached the formation control problem using various methods, such as the virtual structure method \cite{virtual_1, virtual_2, virtual_3}, the behavior-based method \cite{Behavior-basedBalch,Xu2014BehaviorBasedFC, behavioral_3}, the graph-based method \cite{graphbase_1,graphbase_2}, the artificial potential method \cite{apbase_1, apbase_2}, and the leader-follower method \cite{leaderfollowerDas, leadervidal, leaderluca, leaderbesse, leaderdehghani, leaderpedro, leaderyang, leadersakai}. Among these approaches, the leader-follower method has gained widespread popularity owing to its simplicity. This technique involves designating one robot as the leader, which adheres to a predefined trajectory, while other robots act as followers, aligning themselves relative to the leader while adhering to specific inter-robot distance constraints. Notably, the formation problem is simplified to a trajectory tracking problem, with individual robots' control laws ensuring internal formation stability. In linear environments with communication, followers adjust motion based on shared leader data for alignment, assuming wireless data transmission and unlimited sensing range.

In this research, we delve into the leader-follower framework, focusing on two critical aspects that have been overlooked in existing literature. Firstly, we explore the presence of an unknown nonlinear manifold, which introduces a new layer of complexity to the formation control problem. Secondly, we consider the practical limitation of a limited follower sensing range, presenting challenges in effectively localizing the leader for the followers. The primary focus of our research lies in developing a reliable prediction mechanism for the follower robot to anticipate the leader's position effectively. This mechanism ensures that the follower stays within its sensing range, enabling continuous monitoring and maintenance of the desired formation.

This paper is structured as follows: In Section \ref{Methodology}, we introduce our proposed prediction algorithm and outline the experimental setups. Section \ref{Experiment and Discussion} is dedicated to presenting and discussing the experimental results involving mobile robots. Finally, Section \ref{Conclusion} concludes the paper.

\section{Methodology}
\label{Methodology}
Conventional studies on formation control often rely on precise information regarding leader-follower relative positions, velocities, and accelerations to design stable continuous control systems for the follower \cite{leaderfollowerDas, leadervidal, leaderbesse, leaderdehghani}. In linear environments with feasible communication between robots, such as flat terrains with wireless networks, the leader's position and velocity information can be easily shared with followers. Based on this data, each follower calculates its displacement relative to the leader and adjusts its motion accordingly, ensuring alignment and maintaining the desired distance. However, this approach requires either the ability to transmit data wirelessly or unlimited sensing range for the follower to observe the leader continuously.

\begin{figure}[!ht]
  \centering
  \includegraphics[width=8cm]{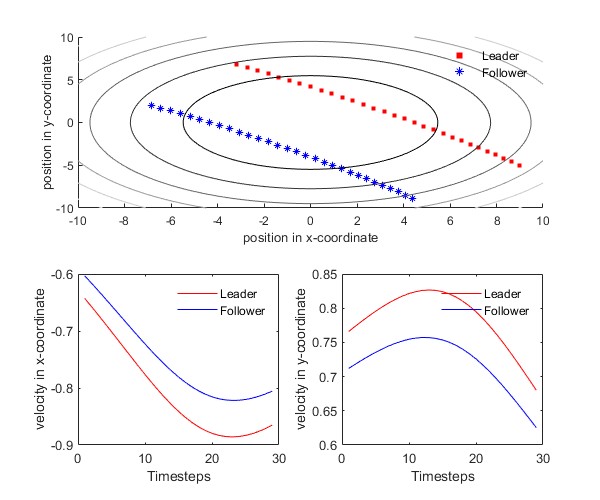}
  \caption{Illustration of leader and follower trajectory in an elliptical paraboloid manifold. Leader robot is labeled in red and follower robot in blue.}
  \label{velocity_compare}
\end{figure}

In contrast, our experiments present two-fold challenges. Firstly, the presence of an unknown nonlinear manifold necessitates maintaining a constant 2D Euclidean distance between the leader and the follower, resulting in distinct velocities for each robot, as shown in Fig. \ref{velocity_compare}. As a consequence, information about the leader's velocity and acceleration becomes irrelevant for our approach.

\begin{figure}[!ht]
        \centering
        \includegraphics[width=8cm]{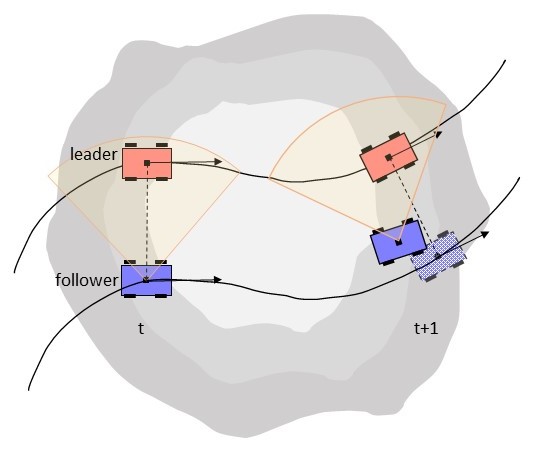}
        \caption{Illustration of leader and follower robots traversing a nonlinear manifold from an overhead view. The leader robot is colored red, while the follower robot is colored blue. The orange area depicts the follower's sensing range. The patterned blue color indicates the ideal position of the follower at timestep $t+1$, while the solid blue color represents the predicted position of the follower at timestep $t+1$.}
        \label{2d_illustration}
\end{figure}

Secondly, the follower robot relies on a visual camera to observe its relative position to the leader, as shown in Fig. \ref{2d_illustration}. However, this camera has a limited field of vision, constraining the follower's sensing range. Consequently, to stay within this limited sensing range, the follower must proactively predict the leader's movements.

To achieve this objective, we present a predictive framework utilizing Koopman operator theory and Extended Dynamic Mode Decomposition (EDMD). By leveraging historical data concerning the follower's movement distances in both the x and y coordinates within its reference frame, we can estimate the corresponding x and y distances the follower will traverse in the subsequent timestep. This enables the follower to steadily approach its intended position, ensuring that the leader remains within the follower's sensing range. Although the predictions might not precisely align with the desired position, the follower's visual camera will serve as the ultimate tool to make minor adjustments in formation alignment. 

\subsection{Koopman operator theory and EDMD}
Bernard O. Koopman's work has established the potential for representing a nonlinear dynamical system using an infinite-dimensional linear operator that operates on a Hilbert space of measurement functions derived from the system's state. The fundamentals of Koopman spectral analysis will now be explored, as discussed in \cite{Koopmanbook, Yoshihiko, mezicpaper}.

Let $g: \mathbf{M} \longrightarrow \mathbb{R}$ be a real-valued measurement function, commonly referred to as \textit{observables}, residing within the infinite-dimensional Hilbert space. The Koopman operator $\mathcal{K}_t$ acts on the observable $g$ as follows:
\begin{align}
\mathcal{K}_tg=g \circ \mathbf{F}_t,
\end{align}
where $\mathbf{F}_t$ is the system dynamic, and $\circ$ is the composition operator.
For discrete-time system with timestep $\Delta t$, it becomes
\begin{align}
 g(\mathbf{x}_{k+1})= \mathcal{K}_{\Delta t}g(\mathbf{x}_k).
\end{align}
While the Koopman operator is linear, it is crucial to recognize that it operates in an infinite-dimensional space. Therefore, it becomes essential to identify significant measurement functions that exhibit linear evolution with the dynamic flow of the system. By performing an eigen-decomposition of the Koopman operator, we can extract a set of measurement functions that effectively capture the system's dynamics while displaying linear behavior over time. A discrete-time Koopman eigenfunction $\varphi(\mathbf{x})$ and its corresponding eigenvalue $\lambda$ satisfies
\begin{align}
 \varphi(\mathbf{x}_{k+1})=\mathcal{K}_{\Delta t}\varphi(\mathbf{x}_k)=\lambda\varphi(\mathbf{x}_k).
\end{align}
Nonlinear dynamics become linear in these eigenfunction coordinate.

In a general dynamic system, the measurement functions can be arranged into a vector $\mathbf{g}$:
\begin{align}
 \mathbf{g}(\mathbf{x})=\begin{bmatrix}
  g_1(\mathbf{x}) \\
  g_2(\mathbf{x}) \\
  \vdots          \\
  g_m(\mathbf{x}) \\
 \end{bmatrix}.
\end{align}
Each measurement functions may be expanded in terms of eigenfunctions $\varphi_j(\mathbf{x})$, thus vector $\mathbf{g}$ can be written as:
\begin{align}
 \mathbf{g}(\mathbf{x})=\sum_{j=1}^{\infty} \varphi_j(\mathbf{x})\mathbf{v}_j,
\end{align}
where $\mathbf{v}_j$ is the $j$-th Koopman mode associated with the eigenfunction $\varphi_j$.
Given this decomposition, we can represent the dynamics of the system in terms of measurement function $\mathbf{g}$ as
\begin{align}
 \begin{split}
  \mathbf{g}(\mathbf{x}_k)&=\mathcal{K}^k_{\Delta t}\mathbf{g}(\mathbf{x}_0)  \\
  &=\mathcal{K}^k_{\Delta t}\sum_{j=0}^{\infty} \varphi_j(\mathbf{x}_0)\mathbf{v}_j \\
  & =\sum_{j=0}^{\infty} K^k_{\Delta t}\varphi_j(\mathbf{x}_0)\mathbf{v}_j \\
  & =\sum_{j=0}^{\infty} \lambda^k_j\varphi_j(\mathbf{x}_0)\mathbf{v}_j.
 \end{split}
\end{align}
The sequence of triples $\{(\lambda_j, \varphi_j, \mathbf{v}_j )\}^{\infty}_{j=0}$ is the Koopman mode decomposition.

Finding such Koopman mode is extremely diffcult even for system with known governing equations. In scenarios where the governing equation is unknown, as is the case in our situation, we turn to the extended dynamic mode decomposition (EDMD) algorithm as proposed by \cite{EDMD_mezic,EDMD_rowley}. EDMD stands as a data-driven technique capable of approximating the Koopman operator without the need for explicit knowledge with the system's governing equations.

We first consider a data set of snapshot pairs $\{(\mathbf{x}_t,\mathbf{y}_t)\}^m_{t=1}$, where $\mathbf{x}_t \in \mathbf{M}$ and $\mathbf{y}_t \in \mathbf{M}$ are snapshots of the dynamic system with $\mathbf{y}_t = \mathbf{F}(\mathbf{x}_t)$, and arrange them into two matrices as
\begin{align}
 \mathbf{X} & =\begin{bmatrix}
  \vert        & \vert        &       & \vert            \\
  \mathbf{x}_1 & \mathbf{x}_2 & \dots & \mathbf{x}_{m} \\
  \vert        & \vert        &       & \vert            \\
 \end{bmatrix},
 \\
 \mathbf{Y} & =\begin{bmatrix}
  \vert        & \vert        &       & \vert          \\
  \mathbf{y}_1 & \mathbf{y}_2 & \dots & \mathbf{y}_{m} \\
  \vert        & \vert        &       & \vert          \\
 \end{bmatrix}.
\end{align}

Additionally, we have a dictionary of observables, denoted as $\mathbf{D} = { \psi_1, \psi_2, \dots, \psi_k }$, where each $\psi_i: \mathbf{M} \rightarrow \mathbb{R}$ is a function mapping to the real numbers. Consequently, we obtain a vector-valued function $\Psi: \mathbf{M} \rightarrow \mathbb{R}^{1\times k}$ such that
\begin{align}
  \Psi(\mathbf{x})=\begin{bmatrix}
    \psi_1(\mathbf{x}) \\
    \psi_2(\mathbf{x}) \\
   \vdots          \\
   \psi_k(\mathbf{x}) \\
  \end{bmatrix}.
 \end{align}

Now consider the span $U(\Psi) = \text{span}\{\psi_1, \dots, \psi_k\} = \{a^T\Psi:a \in \mathbb{C}^k\}.$ Then, for a function $g = a^T \Psi \in U(\Psi)$, we obtain
\begin{align}
  \mathcal{K}g = a^T\mathcal{K}\Psi = a^T \Psi \circ \mathbf{F}. 
  \label{dictionary_koopman}
\end{align}

A finite dimensional representation of the Koopman operator $\mathcal{K}$ is the matrix $K \in \mathbb{R}^{k \times k}$. Then, for Equation (\ref{dictionary_koopman}) to hold for all $a$, we have
\begin{align}
  K \Psi = \Psi \circ \mathbf{F}.
\end{align}

The computation of $K$ involves solving a minimization problem,
\begin{align}
  K = \argmin_{\tilde{K} \in \mathbb{R}^{k \times k}} J(\tilde{K}) = \sum^{m}_{t = 1} \norm{\Psi(\mathbf{y}_t) - \tilde{K}\Psi (\mathbf{x}_t)}^2
\end{align}

The non-invariance of $U(\Psi)$ under $K$ leads to a least squares problem, and
\begin{align}
  K = \Psi( \mathbf{Y})\Psi( \mathbf{X})^\dagger
\end{align}
with $ \mathbf{X}^{\dagger}$ is the pseudo-inverse of $ \mathbf{X}$.

\subsection{Experimental setups and prediction algorithm}
\label{Experimental setups and prediction algorithm}
\begin{figure}[h!]
  \centering
  \subfloat[]{\label{potential}\includegraphics[width=.33\textwidth,height=3.8cm]{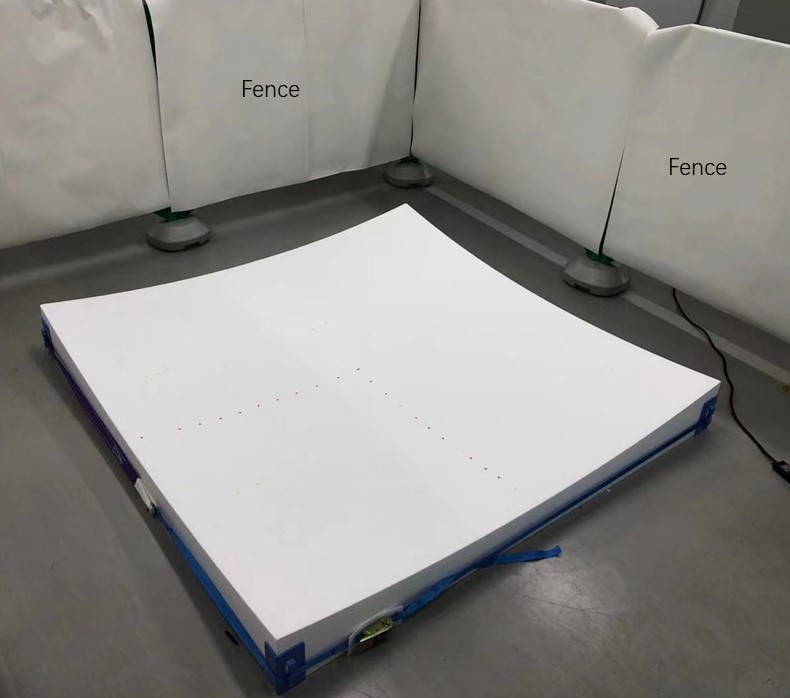}}\hfill
  \subfloat[]{\label{robot_cover}\includegraphics[width=.33\textwidth,height=3.8cm]{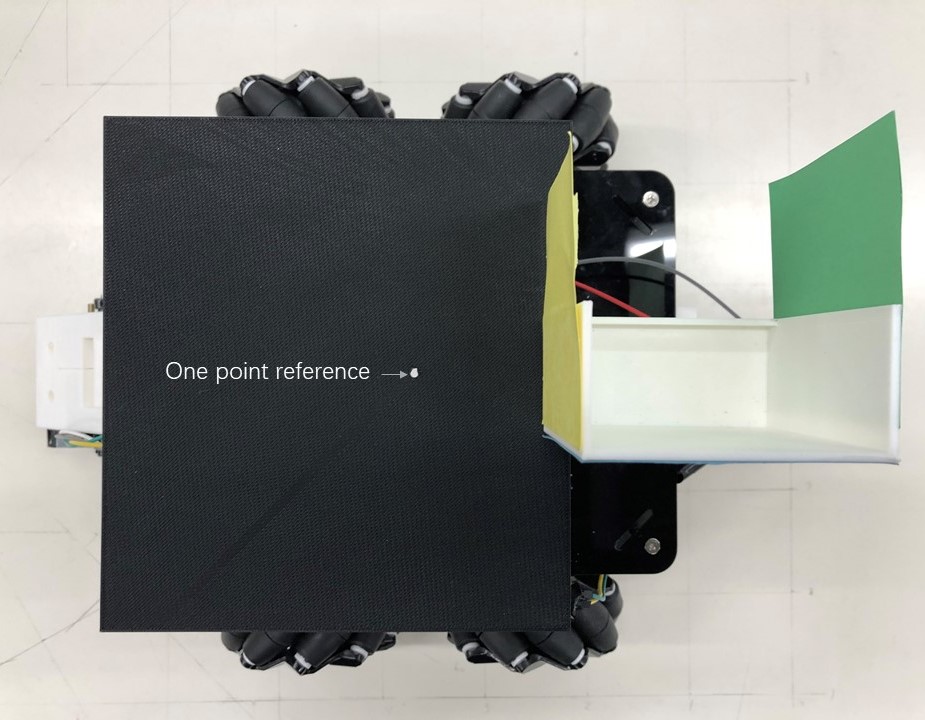}}\hfill
  \subfloat[]{\label{pixy_external_view}\includegraphics[width=.33\textwidth,height=3.8cm]{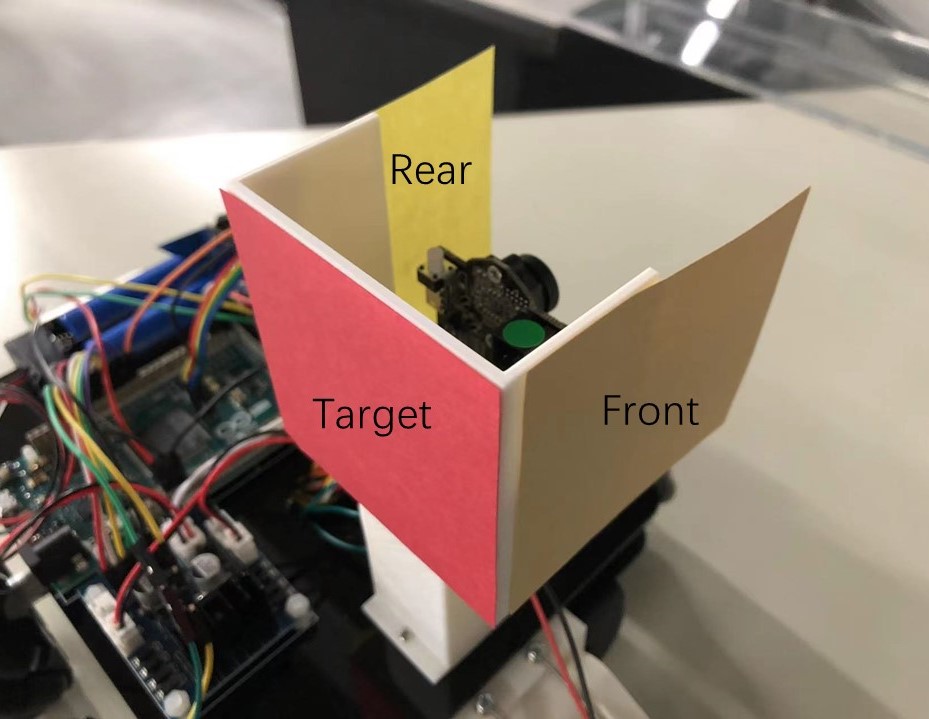}}\hfill\\
  \subfloat[]{\label{encoder_graph}\includegraphics[width=.40\textwidth,height=4.5cm]{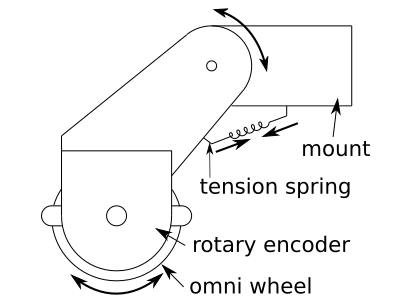}}\hfill
  \subfloat[]{\label{robot_externalview1}\includegraphics[width=.50\textwidth,height=4.8cm]{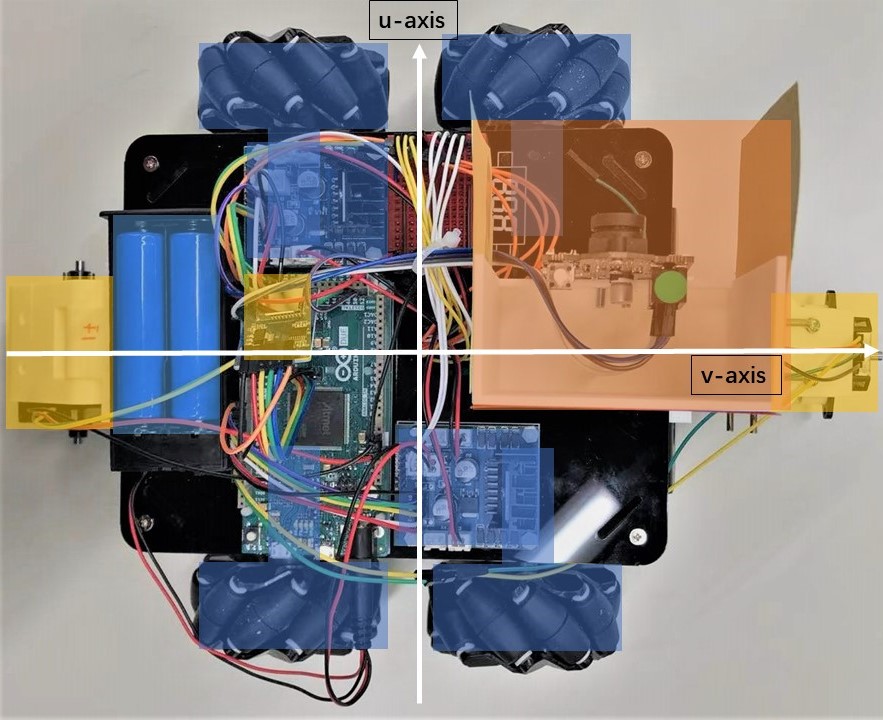}}
  \caption{Experimental setups. (a) Manifold view. The elliptical paraboloid manifold is made of 1780mm by 1780mm polystyrene foam, with minimal height 50mm, and maximum height 208.4mm. (b) Robot with body cover. The white dot is set as the one-point position reference for the robot. (c) View of the robot's visual system. (d) Scheme of robot's sensing system. The sensing system is mounted on the robot's metal skeleton through its supporting platform. The tension spring allows the omni wheel to keep in contact with the manifold surface while travelling. (e) Robot external view. Movement system, sensoring and storage system, and visual system are labeled in blue, yellow and orange, respectively.}
  \label{Experiments}
  \end{figure}

Our experimental setup consists of two omni-directional robots, designated as the leader and the follower, navigating within an elliptical potential, shown in Fig. \ref{potential}. The leader robot maintains a consistent velocity in its self-defined reference frame, achieved through constant wheel rotation for each discrete timestep. Simultaneously, the follower robot endeavors to maintain a fixed separation distance in parallel with the leader, as measured in the Euclidean metric.

To facilitate communication between the two robots, the leader is equipped with a tri-color square marker, serving as a visual reference point for the follower's visual system (Pixy2 camera), as shown in Fig. \ref{pixy_external_view} and Fig. \ref{robot_externalview1}. The follower utilizes this vision system to continuously calculate its relative position with respect to the leader, expressed in terms of angle and distance within the Euclidean metric.

\begin{algorithm}[H]
  \caption{Prediction algorithm}\label{alg:edmd}
  \begin{algorithmic}
  \Require Data set of snapshot pairs $\mathbf{x}^m_{t=1} \equiv {(u_t, v_t)}^m_{t=1}$, with $m = 2$, and $\mathbf{y}_m = \mathbf{x}_{m+1}$
  \Require Iterations for prediction, $M$
  \State $\mathbf{X}_m \gets [\mathbf{x}_1, ... , \mathbf{x}_m]$
  \State $\mathbf{Y}_m \gets [\mathbf{y}_1, ... , \mathbf{y}_m]$
  \State $\mathbf{P}_m \gets (\mathbf{X}_m \mathbf{X}_m^T)^{-1}$
  \State $K_m \gets \mathbf{Y}_m \mathbf{X}_m^{\dagger}$
  \While{$m < M$}
  \State $\mathbf{x}_{m+1} \gets \mathbf{y}_m$   
  \State $\mathbf{y}_{m+1} \gets K_m\mathbf{y}_m$    
  \State $K_{m+1} \gets K_{m}+\frac{(\mathbf{y}_{m+1}-K_{m}\mathbf{x}_{m+1})\mathbf{x}_{m+1}^T\mathbf{P}_{m}}{1+\mathbf{x}_{m+1}^T\mathbf{P}_m\mathbf{x}_{m+1}}$
  \State $\mathbf{P}_{m+1} \gets \mathbf{P}_{m}-\frac{\mathbf{P}_m\mathbf{x}_{m+1}\mathbf{x}_{m+1}^T\mathbf{P}_m}{1+\mathbf{x}_{m+1}^T\mathbf{P}_m\mathbf{x}_{m+1}}$
  \State $m \gets m + 1$
  \EndWhile
  \end{algorithmic}
  \end{algorithm}

A prediction algorithm inspired by both Extended Dynamic Mode Decomposition \cite{EDMD_rowley} and Online Dynamic Mode Decomposition \cite{OnlineDMD} is implemented to enhance the follower's motion control. Algorithm \ref{alg:edmd} shows the pseudo-code for the prediction algorithm. This algorithm utilizes historical data obtained from the follower's own sensing system (refer to Fig. \ref{encoder_graph}), specifically the distance traversed along its u-axis and v-axis. Leveraging this dataset, the prediction algorithm generates one-timestep forecasts for the follower's transverse distances. Remarkably, this prediction framework eliminates the necessity for a global bearing. The algorithm adopts a recursive formulation, minimizing data storage requirements and ensuring computational efficiency by performing matrix inverse calculations only once.

\begin{figure}[!h]
  \centering
  \includegraphics[width = 12cm]{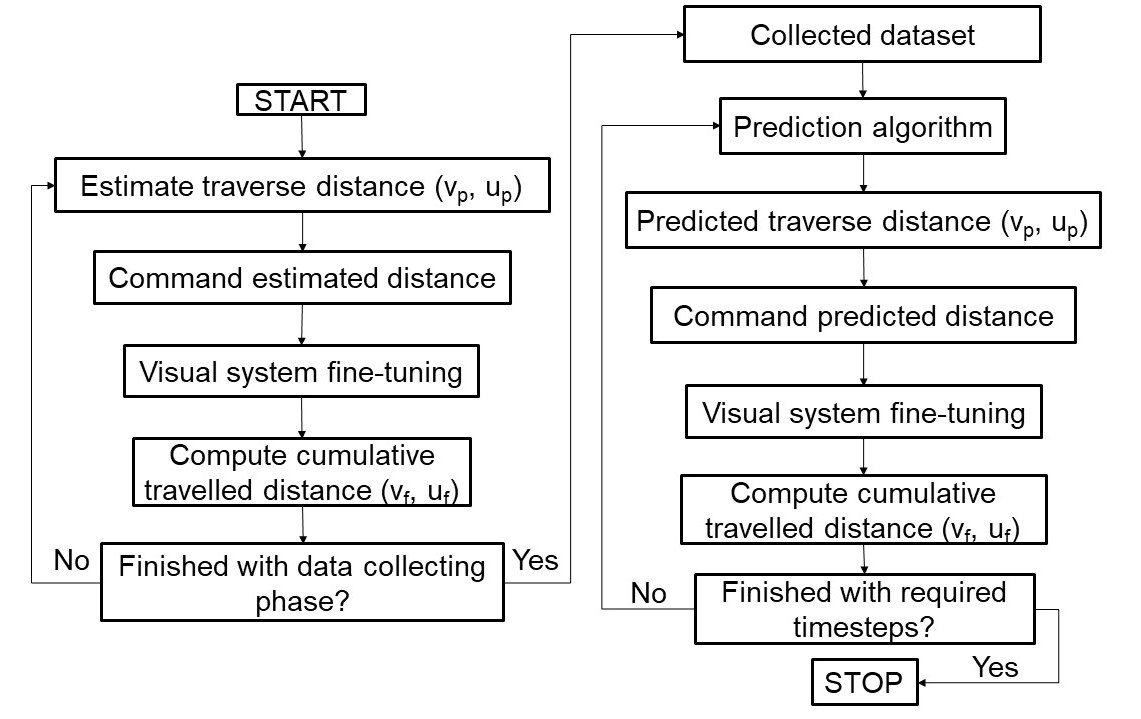}
  \caption{The overall algorithm scheme for follower robot.}
  \label{algorithm_scheme}
 \end{figure}

As shown in Fig. \ref{algorithm_scheme}, the experimental process is structured as follows: during each timestep, the follower employs the prediction algorithm to project its trajectory adjustments. Following the prediction, the follower is directed to traverse a specific distance based on the prediction algorithm's estimation. Subsequently, the follower's position refinement is undertaken through an iterative fine-tuning procedure facilitated by the visual system. The cumulative travelled distance of these iterative fine-tuning adjustments is aggregated to determine the final traversed distance for the respective timestep. Upon achieving the optimal position alignment, the leader robot initiates movement for the subsequent timestep, commencing a new cycle of the process. This sequential pattern of leader movement, follower's prediction and travel, iterative fine-tuning, and distance aggregation is recurrently followed for each timestep.

It is important to note that, before initiating predictions, the prediction algorithm accumulates data over three consecutive timesteps within a designated data collection phase. In this phase, the follower's initial traversed distance aligns with that of the leader, and subsequent traversed distances remain consistent with its own prior timesteps.

To capture the robots' movements comprehensively, an overhead camera takes snapshots of their trajectories during each movement, facilitating trajectory analysis and validation of the algorithm. For more detailed robot's settings and visual fine-tuning process steps please see Appendix.

\section{Experiment and Discussion}
\label{Experiment and Discussion}
\subsection{Experimental results}
\label{Experimental results}
The leader robot is set to travel 5 cm for each given timesteps. The follower robot has an initial velocity equivalent to that of the leader robot, and it uses the algorithms described in section \ref{Experimental setups and prediction algorithm} to control its trajectory.

Two distinct tracks have been selected for the experimentation. The first track, denoted as track one, traverses horizontally on the manifold, spanning a total of 15 timesteps. The second track, referred to as track two, takes a diagonal trajectory and spans a total of 20 timesteps. The initial position of the leader robot is predetermined, while the follower robot is situated at a distance of 32 cm along the v-axis and 0 cm along the u-axis, exhibiting identical alignment relative to the leader robot.

\begin{figure}[H]
  \centering
  \subfloat[][]{\includegraphics[width=2in]{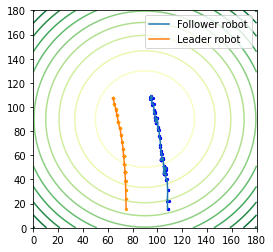}}
  \hfil
  \subfloat[][]{\includegraphics[width=2in]{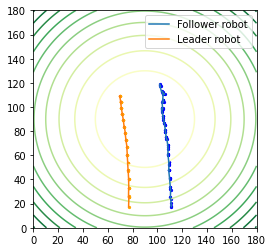}}
 
  \subfloat[][]{\includegraphics[width=2in]{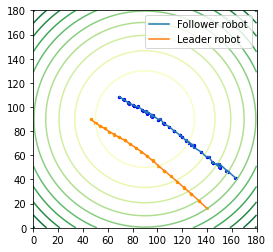}}
  \hfil
  \subfloat[][]{\includegraphics[width=2in]{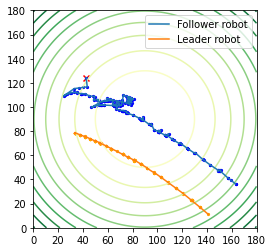}}
  \caption{Trajectory of the leader robot and follower robot in elliptic paraboloid manifold. The trajectory of the leader robot is represented in orange, while the trajectory of the follower robot is depicted in blue. The green contour line in the background represents the manifold. All robots have initial velocity $5$ cm per given timestep. Red cross mark indicates that the follower robot have lost the visual of its beacon neighbour (the leader robot). (a) Track one with prediction algorithm. Initial position of the leader robot is at $(76.3, 15.8)$ cm. Error due to camera distortion is at maximum of 1.3 cm. (b) Track one without prediction algorithm. Initial position of the leader robot is $(76.9, 17.3)$ cm. Error due to camera distortion is at maximum of 1.7 cm. (c) Track two with prediction algorithm. Initial position of the leader robot is $(141.2, 11.1)$ cm. Error due to camera distortion is at maximum of 1.7 cm. (d) Track two without prediction algorithm. Initial position of the leader robot is $(139.7, 16.3)$ cm. Error due to camera distortion is at maximum of 2.7 cm.}
  \label{traj_result}
 \end{figure}

For both tracks, two types of algorithms are employed. Fig. \ref{traj_result} (a) and (c) utilize the prediction algorithm outlined in Algorithm \ref{alg:edmd} to forecast the follower's traverse distances for the upcoming timestep. In contrast, Fig. \ref{traj_result} (b) and (d) employ the traverse distances from the preceding timestep.

Overall, the employed prediction algorithm demonstrated the follower robot's success in effectively tracking the leader robot on both tracks. Notably, in the case of track two, where the prediction algorithm was not utilized, the follower robot lost sight of the leader robot in the final time step.

\begin{figure}[H]
 \centering
 \subfloat[]{\includegraphics[width=0.49\linewidth]{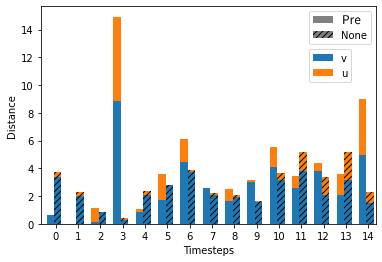}}
 \hfil
 \subfloat[]{\includegraphics[width=0.49\linewidth]{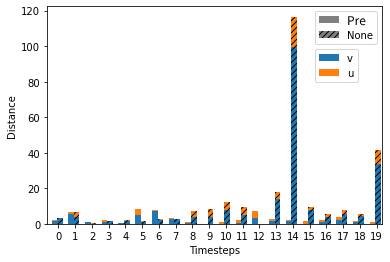}}
 \caption{Correction distance comparison between prediction algorithm and Non-prediciton algorithm for (a) track one and (b) trakc two. For each timestep, distance travelled in v-axis (blue) and u-axis (orange) of follower robot are recorded for both prediction and Non-prediction algorithm (striped).}
 \label{algo_compare}
\end{figure}

To assess the disparity between the follower robot's performance with and without the prediction algorithm, a comparison of correction distances during the visual system fine-tuning stage is presented in Fig. \ref{algo_compare}.

For track one, the correction distance introduced by the fine-tuning process of the visual system exhibits a consistent pattern across most timesteps, regardless of whether the prediction algorithm is utilized. Nevertheless, at timestep 3, the prediction algorithm yields a noticeably larger estimation error when compared to its non-predictive counterpart.

On the other hand, for track two, the integration of the prediction algorithm leads to reduced correction distances for nearly all timesteps, resulting in a significant 78\% overall reduction in correction distances.

\subsection{Discussion}
\subsubsection{On prediction algorithm}
Due to the chosen initial conditions of the leader robot on track one, the curvature of the manifold primarily influences movement along the v-axis of the robot. Additionally, owing to the manifold's symmetry, the trajectory traced by the follower closely mirrors that of the leader. Consequently, as the leader robot's predefined traversal distance remains constant, the follower robot's trajectory closely aligns with minimal variation in traverse distance between each timestep. As a result, employing only the previous timestep's data to guide the follower robot proves as effective as using the prediction algorithm.

The notable error observed in timestep 3 for track one with the prediction algorithm could potentially arise from computational inaccuracies during the calculation of the data matrix's pseudo-inverse. However, it's noteworthy that this error does not escalate in subsequent timesteps, highlighting the robustness of the prediction algorithm against disturbances. 

In contrast, for track two, the leader and follower robots traverse distinct regions of the manifold, and the manifold's curvature exerts differing influences on their trajectories. The utilization of the prediction algorithm leads to improved guidance for the follower robot.

In summary, the effectiveness of the prediction algorithm is demonstrated across the experimental scenarios.

\subsubsection{On limitations}
The objective of the current experiment encompasses the identification of limitations and potential areas for enhancement.
\paragraph{Mobility of robot}
The robot encounters challenges in executing precise movements within regions of the manifold characterized by higher curvature, primarily due to its weight. This limitation adversely affects prediction algorithm, as the presence of significant noise in past data could lead to unreliable predictions. To ameliorate this constraint, potential remedies involve weight reduction measures and the incorporation of PID control mechanisms to enhance the robot's mobility.
\paragraph{Communication delay}
In the context of the experiment outlined in this paper, the leader robot follows a strategy of waiting for the follower robot to attain the desired position before initiating movement at each time step. As the formation's scale increases, this waiting period would proportionally expand, potentially leading to communication delays.
\paragraph{Visual system error}
The experiment relies on the Pixy2 for image recognition within its visual system. The accuracy of the visual fine-tuning process is directly linked to the sensitivity of the Pixy2. The experiment's visual fine-tuning process demands a heightened sensitivity level, which consequently renders it susceptible to image data reading errors caused by flickering. As a future avenue of exploration, alternative sensing methods could be considered to mitigate this visual system error.

\section{Conclusion}
\label{Conclusion}
In conclusion, our research presents an exploration of formation control problem under the leader-follower framework, addressing often-overlooked aspects and validating theoretical formulations through practical experiments with mobile robots. We designed and tested two omnidirectional mobile robots within a nonlinear two-dimensional elliptic paraboloid manifold to validate the formation control algorithm. Notably, the incorporation of the EDMD-based prediction algorithm successfully maintained the desired formation for the follower robot, reducing the workload for position correction by the follower's visual system and enhancing overall formation control efficiency.

Our approach showcases its distinct advantage by effectively tracking and maintaining proximity to the leader, even in the absence of prior leader's velocity and acceleration knowledge. Additionally, the autonomy of our framework, operating independently from data transmission between leader and follower, ensures seamless functionality in environments with restricted or unavailable wireless communication.

Looking forward, opportunities for enhancing performance include refining robot mobility and the visual system. Future work could involve expanding experimentation to larger formations within other types of manifold and validating the formation analysis algorithm as outlined in \cite{Wang} through practical experiments. By bridging theory with tangible results, our research contributes to advancing formation control strategies, offering a pathway to robust and efficient leader-follower systems in real-world scenarios.

% If in two-column mode, this environment will change to single-column format so that long equations can be displayed. 
%\begin{widetext}
%$$\mbox{put long equation here}$$
%\end{widetext}

% Figures should be put into the text as floats. 
% Use the graphics or graphicx packages (distributed with LaTeX2e).
% Put the label that you will use with \ref{} command in the braces of the \label{} command.
%
% \begin{figure}
% \end{figure}

% \begin{table}
% \end{tabular}
% \end{table}

\begin{acknowledgments}
Y.W and T.H acknowledges Professor H. Arai and N. Satoh from Chiba Institute of Technology for helpful discussions. Y.W is supported by the Japanese Government MEXT Scholarship Program.
\end{acknowledgments}

\section*{Data Availability Statement}
The data that support the findings of this study are available from the corresponding author upon reasonable request.

\bibliography{referencer.bib}

\appendix

\section{Manifold specification}
\label{Manifold specification}

Manifold $M$ to be a two-dimensional elliptic paraboloid, which can be explicitly represented as $\frac{1}{40000} (x_1^2+x_2^2) - x_3^2 = 0$.
\begin{figure}[H]
 \centering
 \includegraphics[width=5in]{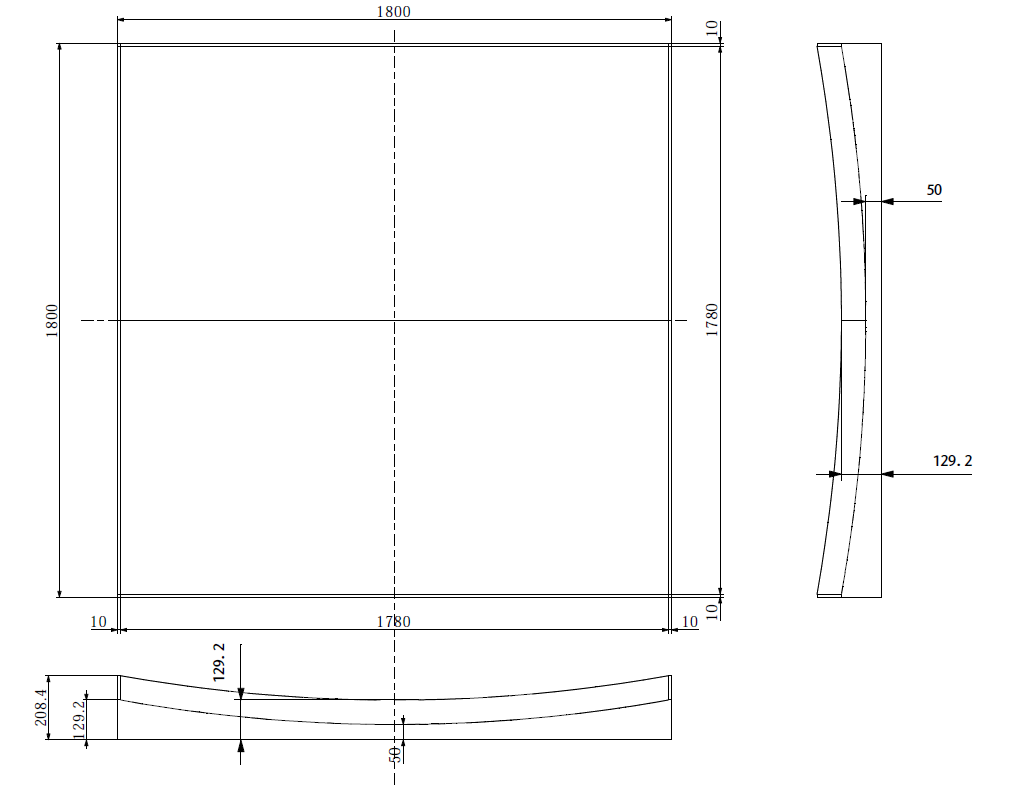}
 \caption{Specification of the manifold. All units in mm.}
 \label{potential_view}
\end{figure}

\section{Design of robot}
\label{Design of robot}
The robot is modified based on OSOYOO\textsuperscript{\tiny\textregistered} Model ZZ012318MC Metal Chassis Mecanum Wheel. It is 365mm in length, 238mm in width, 216mm in height, and weighs 1700 grams. Besides the metal and acrylic skeletons, the robot has three main system: movement system, sensing and storage system, and visual system, all connects to the central processing unit, Arduino Due. 

\subsection{Movement system}
\label{Movement system}
The movement system is responsible for robot's mobility. The system consists of OSOYOO\textsuperscript{\tiny\textregistered} Model-X Motor Driver Module, 18650 battery, DC encoder motor, and Mecanum wheels.
\paragraph{Driver module}
The OSOYOO\textsuperscript{\tiny\textregistered} Model-X motor driver module is an improved L298N module. Two motor driver modules are used to control the front wheels and the rear wheels, respectively.
\paragraph{18650 battery}
The pair of battery acts as the power supply for the whole robot.
\paragraph{DC encoder motor}
There are four DC encoder motors in the robot, corresponding to the four wheels. Each motor consists of two parts: a GM25 DC motor and a dual-channel encoder. The dual-channel encoder can measure wheel rotations of the robot, allowing it to be programmed to travel a preset distance by specifying the number of wheel rotations.
\paragraph{Mecanum wheel}
The Mecanum wheel is an omnidirectional wheel, made with various rubberized rollers obliquely attaching to the wheel rim\cite{wheel}. With a combination of different wheel driving direction, movements to various directions can be performed.

\subsubsection{Movement system calibrations}
\begin{figure}[H]
 \centering
 \subfloat[]{\includegraphics[width=5cm]{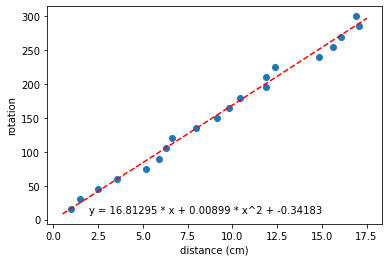}}
 \hfil
 \subfloat[]{\includegraphics[width=5cm]{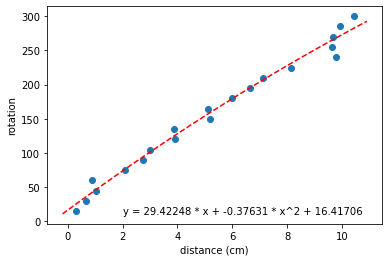}}
 \hfil
 \subfloat[]{\includegraphics[width=5cm]{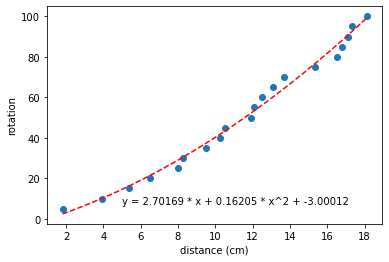}}
 \caption{Motor calibration functions. (a) Movement along v-axis for follower robot. (b) Movement along u-axis for follower robot. (c) Movement along v-axis for leader robot.}
 \label{motor calibration}
\end{figure}

In an ideal condition, all robots would have the same wheel rotation-to-distance conversion function. However, due to the instability and limited accuracy of motors, this conversion function varies for each individual robot. To rectify this, both leader and follower robots are calibrated based on their own distance-to-wheel rotation function as shown in Fig. \ref{motor calibration}.
The motor calibration function is obtained by curve fit the travelled distance measured by observer using one-point reference for various wheel rotations. With the restriction of the size of the manifold and the size of robots using for the experiment, we have focused on finetuning the robots with travelling distance under 20 cm along v-axis, and 10 cm along u-axis. Note the leader robot only calibrated for movement along v-axis, since it only requires to travel in that direction.

\subsection{Sensing system specifications and calibrations}
\begin{figure}[H]
 \centering
 \subfloat[]{\includegraphics[width=7.5cm]{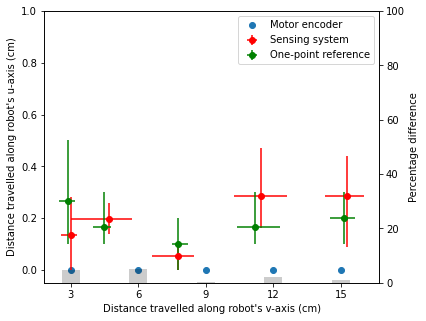}%
  \label{FRW_AC}}
 \hfil
 \subfloat[]{\includegraphics[width=7.5cm]{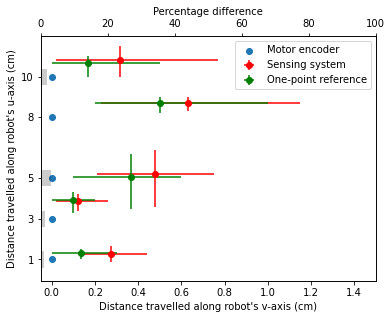}%
  \label{SL_AC}}
 \caption{Travelled distances comparison. Various travelling distances are preset and converted into wheel rotations according to robot's individual motor calibration functions. Travelled distance measured by motor encoder, by sensing system, and by one-point reference method is in blue, red, and green, respectively. Gray bar represents the percentage difference of measured distance between sensing system and one-point reference method. (a) Travelling along v-axis. (b) Travelling along u-axis.}
 \label{encoder_accu}
\end{figure}

The sensing system is implemented for both v-axis and u-axis. It consists of supporting platform, AMT102-V encoder with configuration 512 resolution and 15000 maximum RPM, and 38mm double aluminum omni wheel. Fig. \ref{encoder_accu} demonstrate the travelled distance measured by sensing system, and by observer using one-point reference for different travelling distance.

\subsection{Storage system}
We use a SD card module and 8GB SD card to collect robot's data, such as traversed distances, current timesteps, and number of movements. The storage system is solely for the experiment data analysis. The algorithms of the robots do not require such data collection.

\subsection{Visual system}

\subsubsection{Pixy2 cameara parameters}
\label{Pixy2 parameters}
\begin{table}[H]
 \begin{center}
  \caption{Pixy2 paramter settings\label{paramter table}}
  \begin{tabular}{ |c|c|}
   \hline
   Parameter names           & Values \\
   \hline
   Target signature range    & 5.0    \\
   Front signature range     & 6.0    \\
   Rear signature range      & 6.0    \\
   Camera brightness         & 100    \\
   Block filtering           & 40     \\
   Max merge distance        & 4      \\
   Min block area            & 20     \\
   Signature teach threshold & 5600   \\
   LED brightness            & 4960   \\
   Auto exposure correction  & On     \\
   Auto white balance        & On     \\
   Flicker avoidance         & On     \\
   Mininum frames per second & 30     \\
   \hline
  \end{tabular}
 \end{center}
\end{table}

\subsubsection{Pixy2 camera error}
\label{Pixy2 error}
\begin{table}[H]
 \begin{center}
  \caption{Pixy2 error tolerance for visual algorithm\label{error table}}
  \begin{tabular}{ |c|c|}
   \hline
   Data names             & Tolerance values \\
   \hline
   Side length difference & 4 pixel          \\
   v-axis position        & 5 pixel          \\
   u-axis position        & 1.5 cm           \\
   \hline
  \end{tabular}
 \end{center}
\end{table}

\subsubsection{Visual fine-tuning process calibrations}
The Pixy2 camera detects color blocks and labels them with their respective color signature. 

In order to understand the relationship between Pixy2 image data and robots' relative position and angle, we set the beacon neighbour's position to be at the origin, $(0, 0)$, and use  it as the reference point. We consider a relative distance between the follower robot and its beacon neighbour of $\pm 45$ cm along the u-axis, and $\pm 45$ cm along the v-axis. Image data of Pixy2, including block color signature, block position, and block width and height, are captured every 5 cm. When follower robot's v-axis is parallel to its beacon neighbour, it is considered to have zero relative angle. The relative angle becomes positive in the clockwise direction of the follower robot, with the beacon neighbour taken as the frame of reference. For each position, nine different relative angle, $\pm 60^{\circ}. \pm 45^{\circ}, \pm 30^{\circ}, \pm 15^{\circ}, 0^{\circ}$ are also considered.

\begin{figure}[h!]
 \centering
 \includegraphics[width = 3.5in]{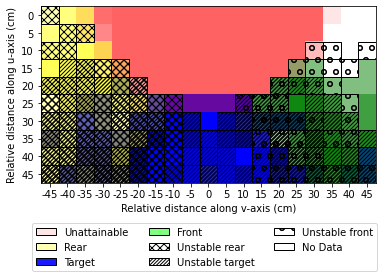}
 \caption{Pixy2 captured images with respect to relative position. All relative angles are overlaid. The overlap between Unattainable area and other areas is due to different relative angles. This overlap do not affects the design of the visual algorithm.}
 \label{overlay_pixy2}
\end{figure}

Based on Fig. \ref{overlay_pixy2}, which is the overlaid image data of all relative angles, we define four areas. The Unattainable area is shown in red, indicating the positions that the follower robot cannot reach due to the volume of robots. The green Front area always captures at least the front color signature block. The blue Target area only captures the target color signature block. The yellow Rear area always captures at least the rear color signature block. The visual algorithm utilizes image data from the Pixy2 camera to determine the movements taken by the follower robot. Its design is intended to guide the follower robot to first reach the Target area and then fine-tune its relative position and angle to achieve the final desired position.

\begin{figure*}[h!]
 \centerline{\includegraphics[width=\textwidth]{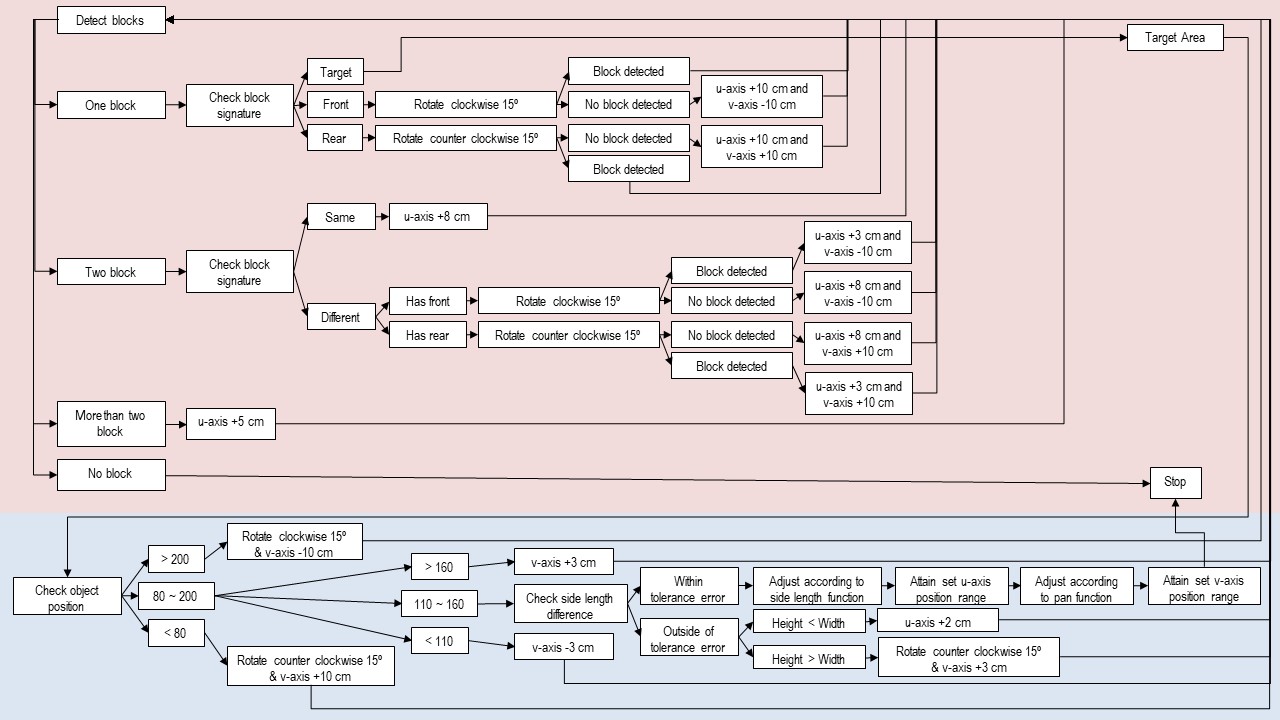}}
 \caption{Visual algorithm flow for entering Target area in purple. Visual algorithm flow for obtaining ideal position while in Target area in red.}
 \label{visual_algo_flow}
\end{figure*}

We briefly summarize the visual algorithm for attaining Target area. The follower robot begins by checking the number of color blocks detected by the Pixy2 camera. Four cases are considered: when one or two blocks are detected, the follower robot checks the color signature of the blocks and moves along the u-axis and v-axis accordingly; when more than two blocks are detected, the follower robot moves along the u-axis and away from its beacon neighbour; when zero blocks are detected, the follower robot stops. Fig. \ref{visual_algo_flow} (in red background) shows the detailed visual algorithm flow. Notice that all movements in the algorithm is taking robot itself as the frame of reference.

\begin{figure}[h!]
 \centering
 \includegraphics[width = 3.5in]{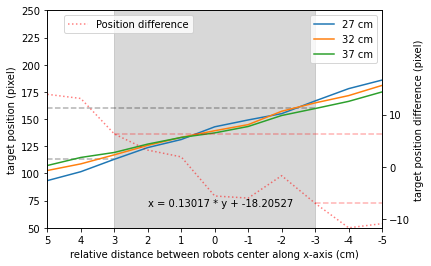}
 \caption{Relative distance along v-axis verse Target color signature block position. Blue, orange, and green lines represent different relative distance between robots along u-axis. The red dotted line represents the maximum difference in Target color signature block position for various relative distance along u-axis. Gray area, $(-3, 3)$ cm represent finetuning area. Inside the finetuning arear, the maximum difference in Target color signature block position for various relative distance along u-axis is less than 7 pixels. The finetuning area corresponds to Target color signature block position range $110 - 160$, shown in horizontal dotted gray lines.}
 \label{pixy_pan}
\end{figure}

\begin{figure}[h!]
 \centering
 \includegraphics[width = 3.2in]{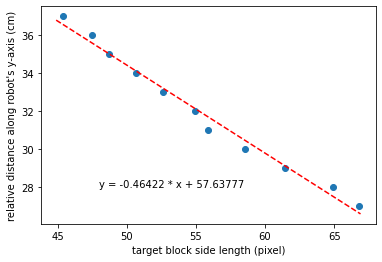}
 \caption{Target block side length verses relative distance along u-axis with zero relative distance along v-axis and zero relative angle.}
 \label{pixy_side}
\end{figure}

Once the follower robot reaches the Target area, fine-tuning movements are necessary. Shown in Fig. \ref{visual_algo_flow} (in blue background), Target color signature block's position, width, and height are used for the finetuning process. We first adjust the position of the Target color signature block, which determines the relative position of the follower robot and its beacon neighbor along the v-axis,  (Fig. \ref{pixy_pan}), then adjust the block's area, which determines the relative distance along u-axis (Fig. \ref{pixy_side}).

Since when adjusting position along v-axis, the follower robot do not have information about its relative position along u-axis, we finetune along v-axis in the range of $(-3, 3)$, where the difference in pan functions between various u-axis distance is small (less than 7 pixel).

\end{document}